%% file: neurips_2025.tex
\documentclass[most]{article}

\usepackage[T1]{fontenc}
\usepackage{pifont}  
\newcommand{\cmark}{\ding{51}}   
\newcommand{\xmark}{\ding{55}} 
\usepackage[preprint]{neurips_2025}

\usepackage[T1]{fontenc}    
\usepackage{hyperref}       
\usepackage{url}            
\usepackage{booktabs}       
\usepackage{amsfonts}       
\usepackage{nicefrac}       
\usepackage{microtype}      
\usepackage{xcolor}         
\usepackage{graphicx}
\usepackage{amsmath}
\usepackage{multirow}
\usepackage{multicol}
\usepackage{colortbl}
\usepackage{tcolorbox}
\usepackage{enumitem}
\usepackage[numbers]{natbib}
\usepackage{tabularx}
\usepackage{subcaption}
\usepackage[T1]{fontenc}    
\usepackage{hyperref}       
\usepackage{url}            
\usepackage{booktabs}       
\usepackage{amsfonts}       
\usepackage{nicefrac}       
\usepackage{microtype}      
\usepackage{xcolor}         
\usepackage{graphicx}
\usepackage{amsmath}
\usepackage{multirow}
\usepackage{multicol}
\usepackage{colortbl}
\usepackage{subcaption}

\usepackage{pifont}    
      
\usepackage{tcolorbox}
\definecolor{light-gray}{gray}{0.6}
\definecolor{front-color}{HTML}{F5FFFA}
\definecolor{Gray}{gray}{0.93}

\usepackage{amssymb}    
\usepackage{pifont} 
   
\definecolor{light-gray}{gray}{0.6}
\definecolor{front-color}{HTML}{F5FFFA}
\definecolor{Gray}{gray}{0.93}
\newtcolorbox{grounderbox}{%
  colback=blue!5, colframe=blue!60!black,
  title=\bfseries Prompt for Grounder,
  fonttitle=\large,
  sharp corners, boxrule=0.8pt, breakable,
  halign title=flush center,
  listing only,
  listing options={basicstyle=\ttfamily\small,breaklines=true}
}

\newtcolorbox{gqabox}{%
  colback=green!5, colframe=green!60!black,
  title=\bfseries Prompt for GQA Agent,
  fonttitle=\large,
  sharp corners, boxrule=0.8pt, breakable,
  halign title=flush center,
  listing only,
  listing options={basicstyle=\ttfamily\small,breaklines=true}
}

\newtcolorbox{verifierbox}{%
  colback=red!5, colframe=red!60!black,
  title=\bfseries Prompt for Verifier,
  fonttitle=\large,
  sharp corners, boxrule=0.8pt, breakable,
  halign title=flush center,
  listing only,
  listing options={basicstyle=\ttfamily\small,breaklines=true}
}

\usepackage[ruled,vlined]{algorithm2e}

\definecolor{light-gray}{gray}{0.6}
\definecolor{front-color}{HTML}{F5FFFA}
\definecolor{Gray}{gray}{0.93}

\title{MUPA: Towards Multi-Path Agentic Reasoning for Grounded Video  Question Answering}

\author{%
  Jisheng Dang\\Lanzhou University, 
  National University of Singapore \\ (jishengdang@gmail.com)
  \And Huilin Song\\
  Sun Yat-sen University \\ (songhlin5@mail2.sysu.edu.cn)
  \And Junbin Xiao\thanks{Corresponding Author}\\
  National University of Singapore \\(junbin@comp.nus.edu.sg)
  \And Bimei Wang\\
  Jinan University
  \AND
  Han Peng\\
  Nanyang Technological University
  \And Haoxuan Li\\
  Peking University
  \And Xun Yang\\
  University of Science and Technology of China
  \And Meng Wang\\
  Hefei University of Technology
  \And Tat-Seng Chua\\
  National University of Singapore
}

\begin{document}

\maketitle

\begin{abstract}
Grounded Video Question Answering (Grounded VideoQA) requires aligning textual answers with explicit visual evidence. However, modern multimodal models often rely on linguistic priors and spurious correlations, resulting in poorly grounded predictions. In this work, we propose MUPA, a cooperative \underline{MU}lti-\underline{P}ath \underline{A}gentic approach that unifies video grounding, question answering, answer reflection and aggregation to tackle Grounded VideoQA. MUPA features three distinct reasoning paths on the interplay of grounding and QA agents in different chronological orders, along with a dedicated reflection agent to judge and aggregate the multi-path results to accomplish consistent QA and grounding. This design markedly improves grounding fidelity without sacrificing answer accuracy. Despite using only 2B parameters, our method outperforms all 7B-scale competitors. When scaled to 7B parameters, MUPA establishes new state-of-the-art results, with Acc@GQA of 30.3\% and 47.4\% on NExT-GQA and DeVE-QA respectively, demonstrating MUPA' effectiveness towards trustworthy video-language understanding. Our code is available in \url{https://github.com/longmalongma/MUPA}.
\end{abstract}

\section{Introduction}
Video Question Answering (VideoQA) has become a key benchmark for evaluating Vision-Language Models (VLMs), as it demands both temporal understanding of dynamic scenes and the ability to generate accurate, open-ended responses. Beyond answer accuracy, an ideal VideoQA system must \emph{visually ground} its predictions in specific video moments, which is a capability essential for interpretability and trust. 
However, recent studies~\cite{liu2025videomind,hu2022lora} reveal a troubling gap: the state-of-the-art (SOTA) VideoQA models, despite their high answer accuracy, often rely on linguistic biases or spurious correlations rather than genuine visual reasoning~\cite{lei2022revealing,wang2024qwen2,xiao2022rethinking,xiao2024can,dang2025synpo,dang2025reinforcing}. For example, while a top-performing model achieves 69\% QA accuracy, 
it correctly grounds only 16\% of its answers, far below the 82\% human level~\cite{xiao2024can}. These findings highlight the urgent need for methods that reliably link answers to visual evidence.

Robust visual grounding remains a fundamental challenge in VideoQA. Existing pipelines predominantly follow a \emph{single-path} or \emph{single-agent} paradigm, where monolithic models map video–question pairs directly to answers without explicit evidence validation. 
While such end-to-end systems (e.g., \emph{TVQA}~\cite{lei2018tvqa} and \emph{FrozenBiLM}~\cite{yang2022zero}) achieve impressive performance on question answering benchmarks, their lack of transparency obscures \emph{why} answers are selected. 
This opacity allows for shortcut learning~\cite{bleeker2024demonstrating}, where models exploit statistical artifacts rather than developing a true understanding. 
From earlier multimodal attention networks to recent transformer-based VLMs, most methods operate as black boxes, lacking mechanisms to verify evidence, resulting in brittle performance and limited trustworthiness.

\begin{figure}[!htbp]
  \centering
  \begin{subfigure}[b]{0.26\textwidth}
    \centering
    \includegraphics[width=\textwidth,height=1.5in]{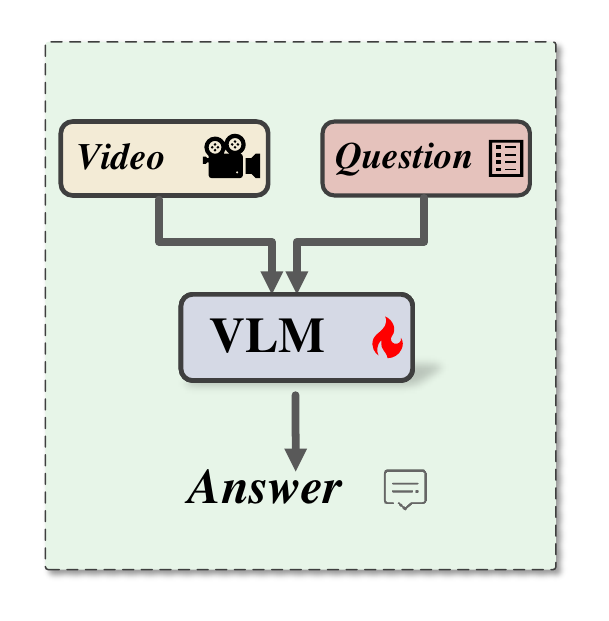}
    \caption{End-to-end approach}
    \label{fig:fig1_a}
  \end{subfigure}
  \hfill
  \begin{subfigure}[b]{0.70\textwidth}
    \centering
    \includegraphics[width=\textwidth,height=1.5in]{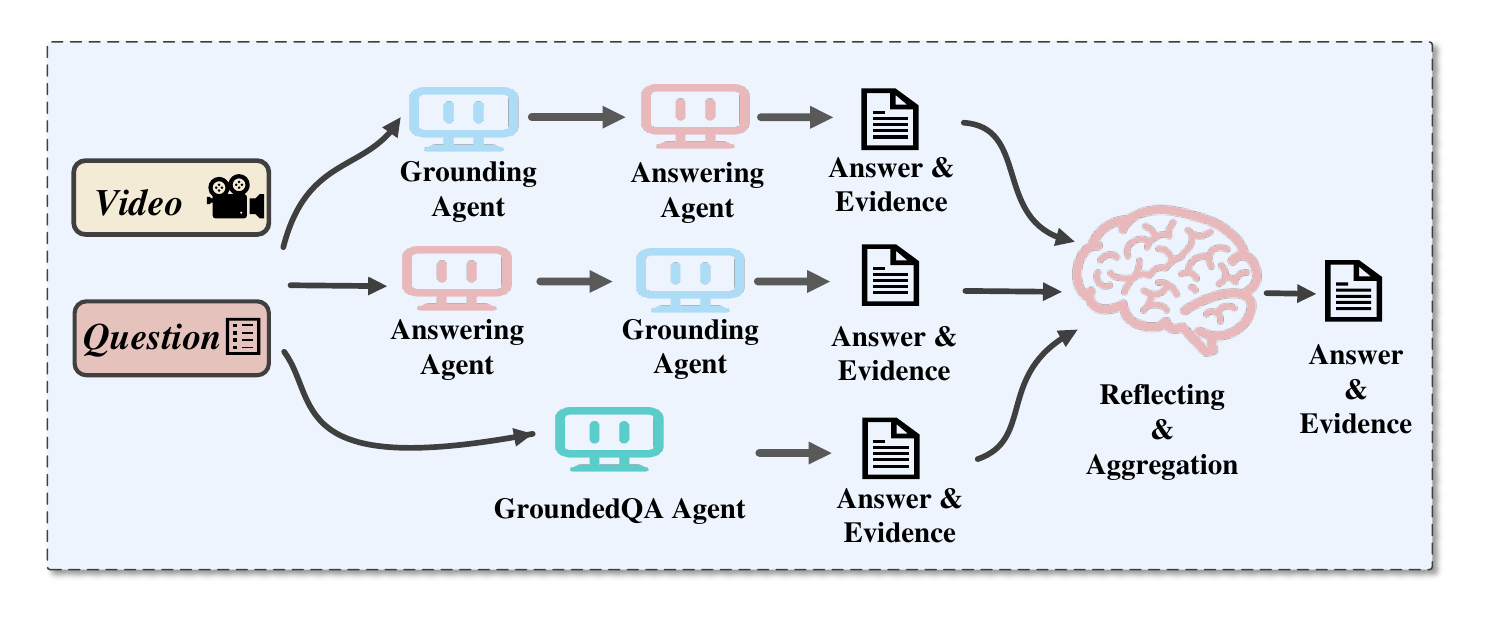}
    \caption{Our multi-agent \& multi-path approach}
    \label{fig:fig1_b}
  \end{subfigure}
\caption{Comparison of end-to-end approach and our multi-agent \& multi-path approach.}
  \label{fig:fig1}
\vspace{-6pt}
\end{figure}

To tackle this challenge, we introduce MUPA, a principled \emph{multi-agent collaborative, multi-path reasoning} framework tailored for Grounded Video Question Answering (Grounded VideoQA). Rather than relying on a single reasoning trajectory, MUPA orchestrates four agents: \textbf{Grounder}, \textbf{Answerer}, \textbf{GQA}, and \textbf{Reflective Agent}. 
The Grounder localizes candidate evidence moments, and the Answerer predicts answers, which collaborate along two reasoning paths (evidence-to-answer and answer-to-evidence). The GQA agent performs grounded question answering through a separate path. 
Each path generates candidate pairs independently, while each pair consists of a proposed answer and its supporting evidence.
Once the three independent reasoning paths have produced their hypothetical answer-evidence candidates, our Reflective Agent engages in a two-stage process, including verification and fusion. In the verification stage, an internal Verifier assesses the coherence between each answer and its evidence, determining whether the evidence truly supports its answer. The Verifier then applies a \emph{product-of-experts} (PoE) mechanism to downweight inconsistent pairs, suppressing predictions where there is weak coherence between the answer and its supporting evidence. In the fusion stage, the agent selects and combines more reliable candidates across paths through a \emph{mixture-of-experts} (MoE) strategy, delivering the final answer and its grounded evidence.
By enforcing the consistency of the cross-agents, MUPA achieves a substantially stronger alignment between the answers and the visual evidence.

We extensively evaluate MUPA on two challenging benchmarks for \emph{Grounded VideoQA}. Qualitatively, MUPA generates answers with precise visual grounding, aligning predictions closely with ground-truth moments. 
Quantitatively, MUPA improves Intersection-over-Prediction (IoP) metrics for visual evidence localization while achieving SOTA performance on \textsc{Next-GQA}~\cite{xiao2024can} and \textsc{DeVE-QA}~\cite{qin2024question}.

Our contributions are threefold:
\begin{itemize}[leftmargin=*]
    \item We propose a collaborative multi-path agentic (MUPA) approach for Grounded VideoQA. MUPA coordinates four specialized agents (Grounder, Answerer, GQA, and Reflectioner) along predefined reasoning paths, effectively unifying VideoQA, Moment Retrieval (MR), and Grounded VideoQA tasks into a compact system that leverages the complementary strengths of each other.
    \item We introduce the reflection mechanism to evaluates the consistency between evidence and answers. It employs PoE to integrate path-level and reflection consistency scores for individual reasoning trajectories, while MoE aggregates multiple candidate answer-evidence pairs through confidence-weighted clustering, significantly reducing reasoning inconsistencies.
    \item We achieve SOTA Grounded QA performance on NExT-GQA (Acc@GQA 30.3\%) and DeVE-QA (Acc@GQA 47.4\%) towards evidenced long and dense video understanding.
\end{itemize}

\section{Related Work}


\subsection{Grounded Video Question Answering}
Early benchmarks such as TGIF-QA~\cite{jang2017tgif}, MSVD-QA~\cite{xu2017video}, ActivityNet-QA~\cite{yu2019activitynet} and NExT-GQA \cite{xiao2021next} advance VideoQA research but do not require models to justify their predictions.  Subsequent analyses reveal that these accuracy-only settings encourage shortcuts rather than genuine video reasoning~\cite{xiao2024can, xiao2025videoqa}. To enforce faithful grounding, datasets like \textsc{NExT-GQA}~\cite{xiao2024can} and \textsc{DeVE-QA}~\cite{qin2024question} pair each question–answer with moment annotations and evaluate the consistency between QA and evidence grounding via grounded QA accuracy.

Grounded VideoQA solutions have evolved along two primary streams.  
\textbf{Two-stage pipelines} first locate candidate segments before answering questions. Representative examples include the Grounder$\to$Answerer workflow in SeViLA \cite{yu2023self} and VideoMind~\cite{liu2025videomind}, where a dedicated agent predicts time spans before a language head generates the answer.  
In contrast, ~\textbf{End-to-end models} integrate temporal grounding directly into cross-modal Transformers. For example, NExT-GQA~\cite{xiao2024can} employs Gaussian-mask optimization to accomplish grounded VideoQA. Grounded-VideoLLM~\cite{wang2024grounded} and Chrono \cite{meinardus2024surprising} introduces discrete time tokens to enhance Video-LLMs for time reasoning.  
Despite these advances, most approaches rely on a direct video-question to answer mapping, which limits their robustness to spurious correlations between questions and video grounding.  Our MUPA framework addresses this limitation by spawning and fusing three complementary paths (grounding-first, answering-first, simultaneous grounding and answering) through a reflection agent. This multi-path agentic approach substantially boosts evidence grounding without sacrificing answer accuracy.


\subsection{Multi-Agent Reasoning}
Generating multiple reasoning traces and reconciling them can mitigate the ``degeneration of thought'' problem in LLMs, where models fixate on initially chosen reasoning paths that are potentially flawed.  
To bridge this gap, Self-Consistency voting~\cite{wang2022self}, CoT Diversity~\cite{wei2022chain}, and Tree-of-Thought search~\cite{yao2023tree} sample diverse chains, while the RR-MP framework~\cite{he2024enhancing} employs \emph{parallel reactive agents} whose outputs are critiqued by a global reflector.  
Orthogonally, ReAct~\cite{yao2023react} and Reflexion~\cite{shinn2023reflexion} interleave acting with self-critique, while debate-style agents~\cite{li2024improving} further enhance robustness.  
Our approach unifies these two research lines: three lightweight reasoning paths (grounding-first, answer-first, retrieval-guided) operate in parallel, after which a Reflective Agent applies PoE and MoE strategies to produce consensus answer-evidence pairs of high mutual support.

\section{Methods}

\begin{figure}[!t]
  \centering
\includegraphics[width=\textwidth]{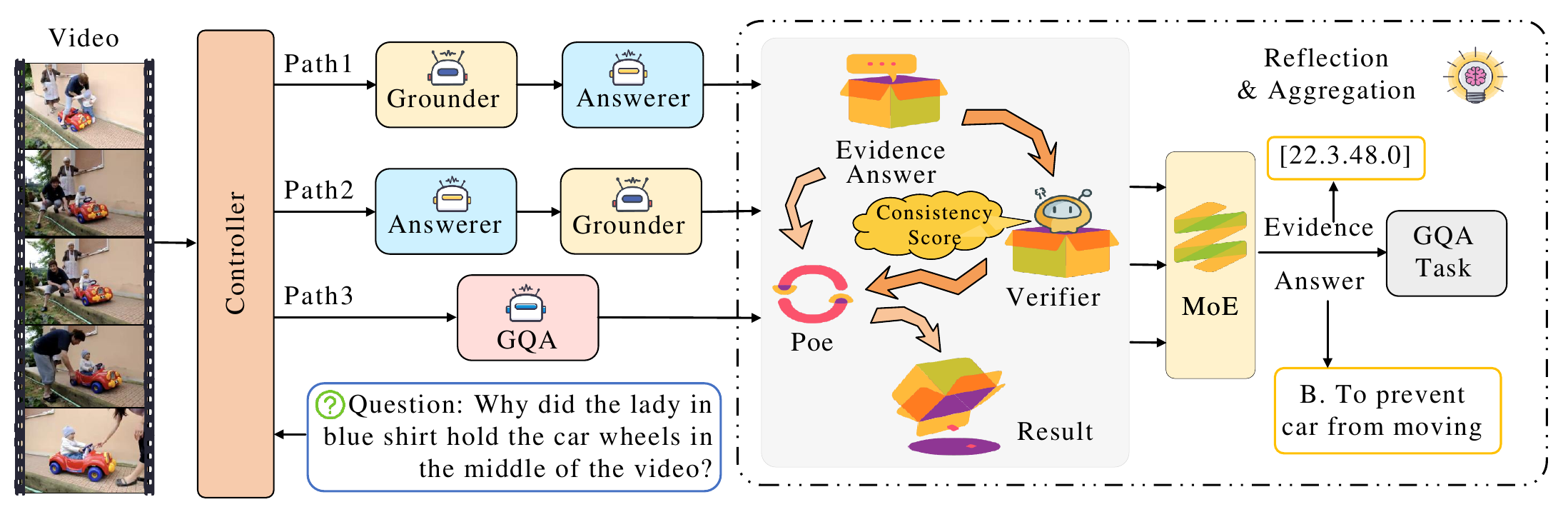} 
\vspace{-12pt}
  \caption{\textbf{Overview.} The controller first senses the incoming task and steers the query to the reasoning routes with multi-path reasoning. Each single-path output is then critiqued by the reflection agent, which employs a verifier followed by a PoE aggregation to refine evidence–answer hypotheses. For GQA, the reflections from Paths 1–3 undergo further reconciliation through a MoE fusion mechanism, yielding  final answer along with its aligned temporal evidence segment.}
  \label{fig:fig2}
  \vspace{-10pt}
\end{figure}

\textbf{Overview.} Figure.~\ref{fig:fig2} illustrates our framework for Grounded VideoQA, which achieves both question answering and precise temporal localization.
Conventional single–agent models often fall back on linguistic shortcuts and yield weak or spurious evidence. 
To address this limitation, we propose MUPA, which fosters reasoning diversity through specialized multi-agent paths and ensures reliability via reflective verification. 
Four collaborative agents power this system: Grounder, Answerer, GQA, and Reflective Agent, each contributing distinct capabilities.
\subsection{MUPA Framework}

\textbf{Multi-path Reasoning.}
MUPA employs three distinct reasoning paths, utilizing a shared pool of three paths to tackle various sub-tasks. 
Paths 1–3 focus specifically on the Grounded VideoQA task, each employing different reasoning strategies:
Path-1 follows a \textit{localize-then-answer} approach, identifying candidate video segments before generating an answer;
Path-2 employs an \textit{answer-then-verify} strategy, presuming an answer before retrieving supporting evidence;
Path-3 utilizes \textit{joint reasoning}, decoding answer–evidence pairs in a single pass under a unified GQA paradigm.
This architecture enables efficient resource sharing while maintaining task-specific reasoning capabilities, thereby keeping the overall model complexity manageable.

\textbf{Reflective Agent for Post-hoc Consolidation.}
The Reflection Agent serves as a critical meta-reasoning component that verifies and integrates the outputs from multiple reasoning paths through two phases. 
First, this agent examines the answer–evidence proposals from each path and assesses their internal consistency before filtering out the implausible or spurious proposals. 
Then, the Reflection Agent consolidates the remaining high-confidence proposals through an aggregation mechanism. The final prediction comprises both the answer and its supporting temporal evidence. 
This post-hoc verification and integration pipeline ensures that the final outputs are reliable and well-grounded, without modifying the internal reasoning processes of the underlying agents.  

\textbf{Independent Training via Chain-of-LoRA.}
Each agent in MUPA is \emph{trained independently} on role-specific data using the Chain-of-LoRA strategy introduced by VideoMind~\cite{liu2025videomind}. This approach allows us to share one frozen vision-language backbone across all agents while attaching lightweight LoRA adapters that are swapped in and out during training and inference for different functional roles.  
By enhancing the model’s localization capabilities, MUPA ensures that answers are driven primarily by video content rather than relying on textual shortcuts. Combining the diversity offered by multiple reasoning paths and the trustworthiness introduced by our post-hoc reflection mechanism, Chain-of-LoRA enables MUPA to produce accurate answers accompanied by precise and defensible temporal evidence. Detailed descriptions of each agent and the reflection mechanism follow in Sections~\ref{sec:multi_path} and~\ref{sec:reflective_agent}, respectively.

\subsection{Multi-Agent Path-Leveraged Evidence Reasoning}\label{sec:multi_path}
Central to MUPA is a suite of independently trained agents, each specialized in a core sub-task: predicting answers, grounding evidence segments, jointly generating answer–evidence pairs, and reflecting the coherence between generated answers and segments. By flexibly composing these agents into three distinct reasoning paths, MUPA automatically addresses Grounded VideoQA, Moment Retrieval, and standard VideoQA tasks. In the following sections, we detail the design and implementation of each agent and reasoning path.

\begin{figure}[t]
  \centering
\includegraphics[width=\textwidth]{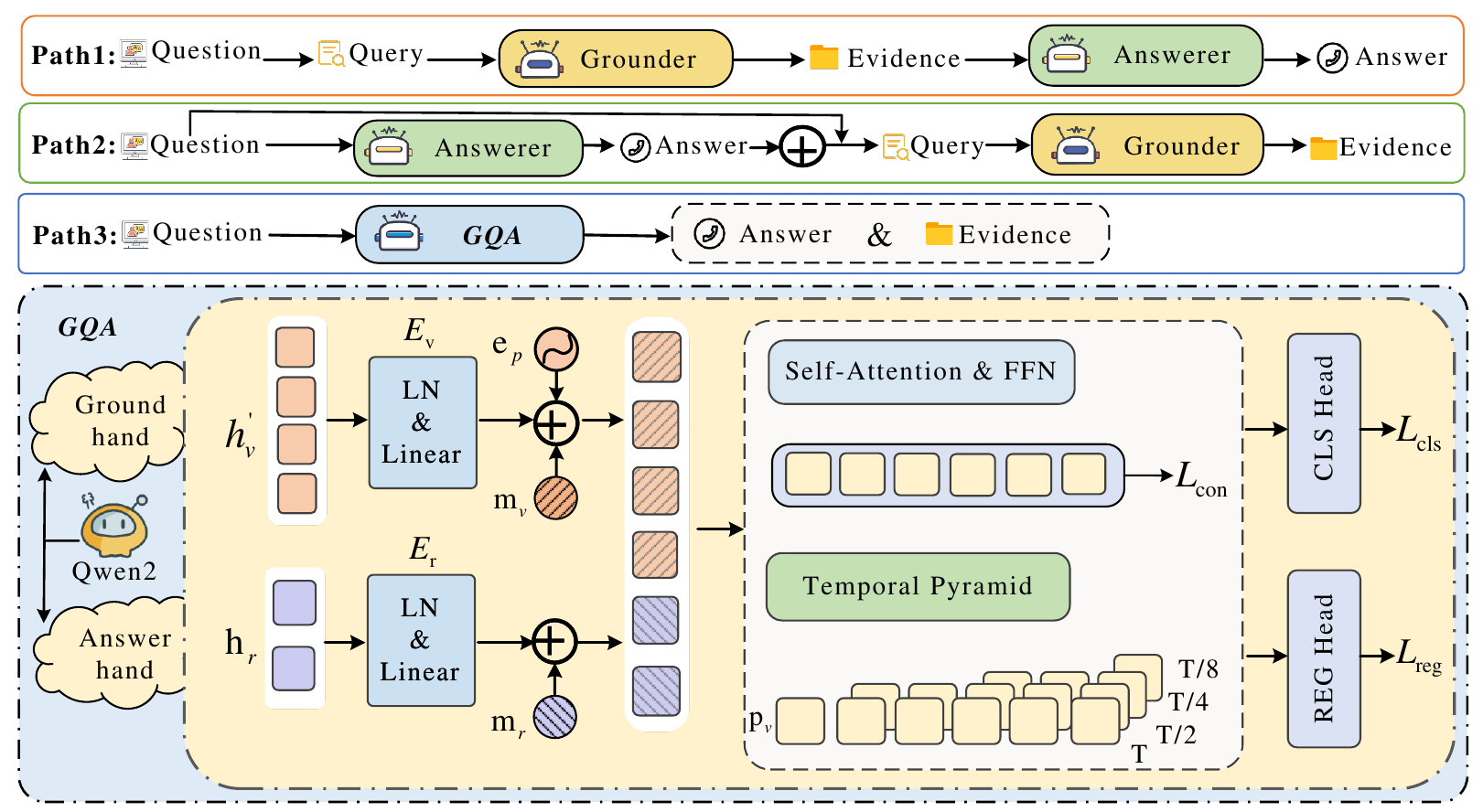} 
\vspace{-8pt}
  \caption{Detailed architecture of multi-path reasoning for Grounded VideoQA. MUPA employs three reasoning paths to generate answer and evidence. Ground head sharing in GQA and Grounder responsible for regressing temporal boundaries}
  \label{fig:fig3}
  \vspace{-20pt}
\end{figure}

\subsubsection{GQA Agent}\label{sec:gqa_agent}

\textbf{Overall Architecture.}
As shown in Figure~\ref{fig:fig3}, our GQA agent integrates a powerful pre-trained cross-modal encoder with two lightweight prediction heads.
We adopt \textbf{Qwen2}~\cite{wang2024qwen2} as the backbone to jointly encode the video frames and the textual question, producing a sequence of multi-modal features that capture both visual and semantic information.
These features are consumed in parallel by  
(i)~an \emph{Answer Head}, which outputs a probability distribution over candidate answers, and  
(ii)~a \emph{Grounder Head}, which predicts the temporal segment(s) that serve as visual evidence.
During training, both the Answer and Grounder heads are optimized simultaneously through a weighted combination of task-specific objectives:
\begin{equation}
\mathcal{L}_{\text{GQA}} = \lambda_{\text{ans}}\mathcal{L}_{\text{ans}} + \lambda_{\text{loc}}\mathcal{L}_{\text{loc}},
\end{equation}
where $\lambda_{\text{ans}}$ and $\lambda_{\text{loc}}$ are the balancing weights. The answer prediction loss $\mathcal{L}_{\text{ans}}$ and the temporal localization loss $\mathcal{L}_{\text{loc}}$ are detailed below.

\textbf{Role within the MUPA.}
The GQA agent constitutes \textbf{Path-3} in MUPA’s design, implementing a joint reasoning approach. Unlike Path 1~2 that follows the sequential reasoning strategy, this single-pass solution generates an answer–evidence pair at the same time.
Its Answer and Grounder heads are also reused either in complementary orders by Path-1 (ground$\to$answer) and Path-2 (answer$\to$ground), illustrating the modularity of our agent-centric approach.

\textbf{Answer Head.}\label{sec:answer_head}
The Answer Head treats Grounded VideoQA as a $K$-way classification problem, where $K$ represents the number of candidate answers provided by the dataset.
A linear projection maps the \texttt{[CLS]}-style token $h_\mathrm{cls}$ to logits $\mathbf z\in\mathbb R^{K}$, followed by a softmax normalization to obtain answer probabilities $\mathbf{p} = [p_1, p_2, \ldots, p_K]$.
We optimize this head by minimizing the standard categorical cross-entropy as follows:
\vspace{-5pt}
\begin{equation}
\mathcal L_{\text{ans}}
\;=\;
-\sum_{c=1}^{K} y_c \log p_c ,
\end{equation}
where $\mathbf{y} = [y_1, y_2, \ldots, y_K]$ is the ground-truth vector.

\textbf{Grounder Head.}\label{sec:grounder_head}
As shown in Figure.~\ref{fig:fig3}, the grounder head follows the VideoMind design~\cite{liu2025videomind}. It begins by applying layer normalization and a linear projection to both the VLM frame embeddings and the \texttt{<REG>} token embedding. Modality embeddings and positional encodings are then added before feature fusion. The fused features are processed in two parallel branches. One branch passes them through a lightweight self‐attention module and a feed‐forward network to produce evidence classification features, which incur the classification loss \(L_{\mathrm{cls}}\). The other branch sends them into a temporal pyramid encoder to aggregate multi‐scale context and then regresses the start and end timestamps via the REG head, yielding the regression loss \(L_{\mathrm{reg}}\). A consistency loss \(L_{\mathrm{con}}\) is also computed on the attention features to enforce alignment. The overall grounding objective is  
\[
\mathcal{L}_{\mathrm{loc}} = L_{\mathrm{cls}} + L_{\mathrm{reg}} + L_{\mathrm{con}}.
\]

\subsubsection{Cascaded Paths: Ground-first vs.\ Answer-first}\label{sec:cascaded_paths}

In addition to the single pass GQA agent, MUPA implements two complementary cascaded trajectories that assemble the independently trained Grounder and Answerer in opposite orders. Each trajectory captures a distinct reasoning pattern.

\textbf{Path-1 (Ground-first-answer-second).} 
As shown in Figure.~\ref{fig:fig3}, Path-1 starts by converting the question into a query: we remove the WH word and any leading auxiliary verb, yielding the template  
\texttt{``The moment when [core clause]''}.  
The Grounder then selects the top \(k\) temporal spans that match this query. Concatenated with the original question, these clips are forwarded to the Answerer. This ordering accords with human intuition and remains the mainstream strategy in current Grounded VideoQA research.

\textbf{Path-2 (Answer-first-ground-second).} 
As shown in Figure.~\ref{fig:fig3}, Path-2 reverses the order. The Answerer first predicts a provisional answer from the full video and question. This answer is then normalized and combined with the core clause of the question to form  
\texttt{``The moment when [core clause] [normalised answer]''}.  
This query is then passed to the Grounder, which retrieves supportive temporal segments. Path-2 thus arises from a reverse-thinking adaptation of Path-1.

\textbf{Why Introducing Answer-first Path?} A query built solely from the question may misalign with the video, so an early grounding error in Path-1 often propagates to the answer. Meanwhile, modern models generally exhibit stronger question-answering capability than temporal localization, even in zero-shot settings. By combining a provisional answer into the query, Path-2 narrows the semantic gap between text and video when the answer is correctly predicted, improving localization accuracy. If the predicted answer is incorrect, this mismatch lowers the confidence of the Grounder, which signals the Reflective Agent to downweight unreliable corresponding evidence. Hence, the two cascaded trajectories provide complementary cues that the reflection stage later reconciles.


\subsection{Reflective Agent}\label{sec:reflective_agent}

The diversity provided by MUPA’s multiple reasoning paths is valuable but still comes at the cost of potential contradictions and occasional hallucinated evidence. 
To safeguard reliability, we introduce a \textbf{Reflection Agent}
that evaluates through the candidate answer–evidence pairs emitted by the lower-level agents and performs an \emph{independent post-hoc appraisal}.  
This agent functions as a meta-reasoner rather than an additional reasoning path, performing two key operations:
(i) scrutinizing the internal coherence of each answer-evidence pair, 
and (ii) integrating consistent findings across different paths into a unified consensus.  
By training the Reflection Agent separately without back-propagating gradients to the underlying agents, we create an impartial auditor that enforces faithfulness while preserving the specialized behaviors developed by individual reasoning paths.

\subsubsection{Single-Path Verification}\label{sec:verifier_poe}

\textbf{Verifier.}\label{sec:verifier}
The Verifier Follows VideoMind design~\cite{liu2025videomind},  evaluating the top-$N$ spans ($N=5$) produced by the Grounder. Each candidate span is temporally extended by 50\,\% and spatially upscaled, and special tokens \texttt{<SEG\_START>} and \texttt{<SEG\_END>} are inserted at its boundaries. The Verifier processes the marked clip together with the question and outputs logits $\ell_{\mathrm{yes}}$ and $\ell_{\mathrm{no}}$. These are converted into a consistency score  
\[
v_{ik} \;=\; \sigma\bigl(\ell_{\mathrm{yes}} - \ell_{\mathrm{no}}\bigr),
\]
where $\sigma$ is the sigmoid function, $k=1,\dots,N$ indexes the spans, and $i$ indexes the reasoning path. During training, spans with $\mathrm{IoU}>0.5$ are treated as positive examples, and the Verifier is fine-tuned via a lightweight LoRA adapter. At inference, the score $v_{ik}$ reflects the confidence that the queried event occurs within the $k$-th span of path $i$.

\textbf{PoE Re-scoring.}  
We denote the original confidence score assigned to span $S_{ik}$ by the Grounder as $c_{ik}$ and the consistency score from the Verifier as $v_{ik}$. 
After interpreting $c_{ik}$ and $v_{ik}$ as independent posterior probabilities, we combine them using the PoE, to obtain the fused confidence score, $p_{ik} \;=\; c_{ik}\,v_{ik}, k=1, 2, \dots,N$.
The spans are re-ranked by $p_{ik}$.
The top one, $\hat S_i = S_{ik^\star}$ with $k^\star=\arg\max_k p_{ik}$, is retained together with the path-level confidence $p_i = p_{ik^\star}$ for the subsequent multi-path fusion.

\textbf{Benefits.}  
Verifier eliminates hallucinated evidence while preserving the original answer predictions, whereas the parameter-free PoE rule promotes a span \emph{only} when both the path and the Verifier are confident, thereby suppressing false positives from either source.
The zoom-in operation sharpens boundary awareness, and explicit boundary tokens raise the model’s sensitivity to precise span limits.  
Moreover, the consistency score places the Verifier’s output on the same probabilistic scale as the Grounder's confidence, which makes the PoE fusion rule both concise and effective.
Together, these design choices yield more reliable and better-localized evidence without introducing any additional trainable parameters to the system.

\subsubsection{Multi-Path Fusion}\label{sec:multi_fusion}

\textbf{Overview.}
After single-path verification, each reasoning path $i$ produces one answer
\(A_i\) together with a set of confidence-weighted spans
\(\{(S_{ik},p_{ik})\}_{k=1}^{N_i}\).
The Reflection Agent first selects a single consensus answer and then merges all candidate spans into the top-\(K\) evidence segments through a Mixture-of-Experts (MoE) procedure that combines confidence weighting, weighted clustering, and boundary refinement.

\textbf{Answer Consolidation.}
To determine the consensus answer, we implement a weighted majority voting mechanism:
\begin{equation}
\hat A=\arg\max_{a}\sum_{i}\sum_{k}p_{ik}\,\mathbf1[A_i=a].
\end{equation}
\textbf{MoE Weighting of Spans.}
For every span, we normalize its confidence according to the weights
\(
w_{ik}=p_{ik}/\sum_{j,\ell}p_{j\ell}
\).
These weights serve as expert reliabilities in the subsequent fusion.

\textbf{Weighted \(k\)-means Clustering.}
To aggregate the candidate pairs generated by the agents, we first map each temporal span \(S_{ik}=(s_{ik},e_{ik})\) from $i$-th reasoning path and $k$-th candidate to the point
\(\mathbf x_{ik}=(s_{ik},e_{ik})\in\mathbb R^{2}\).
To discover up to \(K\) shared moments across different reasoning paths, we formulate a weighted clustering problem:
\begin{equation}
\min_{\{\mathbf C_j\},\,\{c(ik)\}}
\sum_{i,k} w_{ik}\,\bigl\|\mathbf x_{ik}-\mathbf C_{c(ik)}\bigr\|^{2},
\label{eq:wkmeans}
\end{equation}
where \(\mathbf C_j\) represents the centre of cluster \(j\),
\(c(ik)\in\{1, 2, \dots, K\}\) assigns each span to a cluster, and $w_{ik}$ denotes the confidence score associated with each candidate pair.
This optimization in Equation \eqref{eq:wkmeans} minimizes the weighted within-cluster variance and is equivalent to maximizing the likelihood of a Gaussian mixture model with isotropic components under weights \(w_{ik}\).
This clustering approach enables us to identify and achieve the most consistent evidence segments across multiple reasoning trajectories.

\textbf{Boundary Refinement.}
For each cluster $j$, we define the set of indices $I_j=\{(i,k)\mid c(ik)=j\}$ representing all spans assigned to that cluster.
We then refine each centre by solving a weighted least-squares problem formed as follows:
\begin{equation}
\tilde{\mathbf C}_j
=\frac{\sum_{(i,k)\in I_j}w_{ik}\,\mathbf x_{ik}}
       {\sum_{(i,k)\in I_j}w_{ik}},
\end{equation}
which is the Gauss–Markov minimum-variance estimate of the true boundary.
Each refined center \(\tilde{\mathbf C}_j=(\tilde s_j,\tilde e_j)\) corresponds to a temporal span $\tilde S_j = (\tilde s_j, \tilde e_j)$, collectively forming the final evidence set \(\{\tilde S_j\}_{j=1}^{K}\).

The complete fusion module returns the consolidated answer \(\hat A\) along with the top-\(K\) refined evidence spans \(\{\tilde S_j\}\).
This design leverages span-level confidences, suppresses outliers through MoE weighting, and achieves precise boundary estimates with only closed-form computations without introducing any additional trainable parameters.\\

\begin{table}[!htbp]
\vspace{-15pt}
\centering
\caption{Zero-shot results on NExT-GQA~\cite{xiao2024can}. Acc@GQA denotes QA accuracy when IoP$\ge0.5$.}
\small
\begin{tabular}{cc|ccc|ccc|c}
\toprule
\multirow{2}{*}{Method} & \multirow{2}{*}{Size} & \multicolumn{3}{c|}{IoU} & \multicolumn{3}{c|}{IoP} & \multirow{2}{*}{Acc@GQA} \\
\cmidrule(lr){3-5} \cmidrule(lr){6-8}
 & & R@0.3 & R@0.5 & mIoU & R@0.3 & R@0.5 & mIoP & \\
\midrule
FrozenBiLM NG+~\cite{yang2022zero}    & 890M & 13.5 & 6.1  & 9.6  & 28.5 & 23.7 & 24.2 & 17.5 \\
VIOLETv2~\cite{fu2023empirical}     & --   & 4.3  & 1.3  & 3.1  & 25.1 & 23.3 & 23.6 & 12.8 \\
SeViLA~\cite{yu2023self}            & 4B   & 29.2 & 13.8 & 21.7 & 34.7 & 22.9 & 29.5 & 16.6 \\
LangRepo~\cite{kahatapitiya2024language}          & 8×7B & --   & 12.2 & 18.5 & --   & 28.7 & 31.3 & 17.1 \\
VideoStreaming~\cite{qian2024streaming}    & 8.3B & --   & 13.3 & 19.3 & --   & 31.0 & 32.2 & 17.8 \\
LLoVi~\cite{zhang2023simple}             & 1.8T & --   & 15.3 & 20.0 & --   & 36.9 & 37.3 & 24.3 \\
HawkEye~\cite{wang2024hawkeye}           & 7B   & 37.0 & 19.5 & 25.7 & --   & --   & --   & --   \\
VideoChat-TPO~\cite{yan2024task}     & 7B   & 41.2 & 23.4 & 27.7 & 47.5 & 32.8 & 35.6 & 25.5 \\
VideoMind-2B~\cite{liu2025videomind}  & 2B   & 45.2 & 23.2 & 28.6 & 51.3 & 32.6 & 36.4 & 25.2 \\
VideoMind-7B~\cite{liu2025videomind}  & 7B   & 50.2 & 25.8 & 31.4 & 56.0 & 35.3 & 39.0 & 28.2 \\
\midrule
\rowcolor[gray]{0.9}
MUPA-2B (Ours)   & 2B   & 49.4  & 25.6  & 27.2  & 57.0  & 38.7  & 39.1  & 28.7  \\
\rowcolor[gray]{0.9}
MUPA-7B (Ours)   & 7B   & \textbf{54.2}  & \textbf{27.3}  & \textbf{33.4}  & \textbf{60.6}  & \textbf{39.4}  & \textbf{41.4}  & \textbf{30.3}  \\
\bottomrule
\end{tabular}
\label{tab:nextgqa}
\vspace{-20pt}
\end{table}

\section{Experiments}

In this section, we evaluate MUPA on multiple benchmarks, addressing three research questions:
\textbf{RQ1:} Is MUPA competitive with task‑specific baselines?
\textbf{RQ2:} What effects does Multi-path contribute?
\textbf{RQ3:} Does the Reflection Agent boost grounding fidelity and answer accuracy?
Grounded VideoQA requires both the answer prediction and temporal grounding. We evaluate MUPA on two Grounded VideoQA datasets, NExT-GQA~\cite{xiao2024can} and DeVE-QA~\cite{qin2024question}. More experiments and details are in the supplementary.

\begin{table}[!htbp]
\vspace{-5pt}
\centering
\caption{Grounded VideoQA results on DeVE-QA~\cite{qin2024question}. IoP@0.5: rank-1 recalls at threshold 0.5.}
\small
\begin{tabular}{l c c c c c c}
\toprule
Model & mIoP & IoP@0.5 & mIoU & IoU@0.5 & Acc@QA & Acc@GQA \\
\midrule
\multicolumn{7}{l}{\textcolor{gray}{\textit{Weakly-supervised}}} \\
FrozenBiLM(NG+)~\cite{xiao2024can}    & 21.2 & 18.2 & 8.50 & 6.2  & 61.6 & 14.5 \\
Temp[CLIP](NG+)~\cite{xiao2024can}     & 24.6 & 24.8 & 12.5 & 9.1  & 58.9 & 14.9 \\
SeViLA$^{*}$~\cite{yu2023self}         & 25.8 & 19.9 & 21.2 & 11.5 & 62.7 & 16.1 \\
\midrule
\multicolumn{7}{l}{\textcolor{gray}{\textit{Zero-shot}}} \\
LLoVi~\cite{zhang2023simple}                & 27.5 & 27.0 & 17.9 & 12.9 & 63.9 & 22.8 \\
DeVi-Gemini2.0~\cite{qin2024question}       & 36.7 & 32.8 & 22.2 & 18.8 & 72.4 & 27.9 \\
VideoMind-2B~\cite{liu2025videomind}         & 49.9 & 50.7 & 26.3 & 21.7 & 76.5 & 41.2 \\
VideoMind-7B~\cite{liu2025videomind}         & 51.9 & 52.3 &  \textbf{30.1}  &  \textbf{26.5}  &  \textbf{81.0}  & 44.2 \\
\midrule
\rowcolor[gray]{0.9}
MUPA-2B (Ours)      & 52.6 & 53.3 & 27.9 & 24.2 & 76.3 & 43.9 \\
\rowcolor[gray]{0.9}
MUPA-7B (Ours)      &  \textbf{53.9}   &  \textbf{55.2}  & 29.2 & 26.3 & 80.3 &  \textbf{47.4}  \\
\bottomrule
\end{tabular}
\label{tab:deveqa}
\vspace{-15pt}
\end{table}
 
\subsection{Quantitative  Results}

We further evaluate MUPA on several additional benchmarks spanning both grounded VideoQA and Moment Retrieval. Across the board, our multi-path architecture continues to yield strong performance. Experiments on more datasets and benchmarks can be found in the supplementary.

\noindent\textbf{GQA Results.}
Table~\ref{tab:nextgqa} reports zero-shot results on the challenging \textsc{NExT-GQA} benchmark.
MUPA achieves 29.0\% Acc@GQA and 38.7\% IoP\@0.5, outperforming all existing 7B-scale competitors despite using only \textbf{2B} parameters.
Specifically, it surpasses VideoMind-7B, the previous best 7B model, by +0.8 pp (percentage point) Acc@GQA and +3.4 pp IoP\,R@0.5.
Moreover, MUPA demonstrates even more substantial performance gains when compared to VideoChat-TPO and HawkEye.
These improvements remain consistent across all IoP metrics, with \textsc{MUPA-2B} achieving a mIoP of \textbf{39.7\%}, an absolute gain of +3.3 pp over VideoMind-7B and +2.4 pp over the considerably larger LLoVi model with 1.8T parameter count.
When scaled to 7B parameters, MUPA-7B attains 30.3\% Acc@GQA and 39.4\% IoP R@0.5, establishing the current SOTA.
Our evaluation confirms that MUPA effectively balances answer correctness with temporal grounding precision, outperforming all larger alternative models. Without sacrificing QA accuracy, it delivers more precise evidence localization across all IoP metrics, establishing new SOTA performance.
Table~\ref{tab:deveqa} reports zero-shot results on the challenging \textsc{DeVE-QA} benchmark. DeVE-QA emphasizes fine-grained evidence alignment between answers and video spans.


\textbf{Is MUPA Competitive with Task‑specific Baselines?} Across multiple benchmarks, MUPA demonstrates superior performance in both answer accuracy and evidence grounding. On \textsc{NExT‑GQA}, MUPA‑2B already surpasses the best 7B predecessor in both Acc@GQA and IoP@0.5 metrics, while MUPA‑7B sets a new SOTA. Similar trends hold for DeVE‑QA, where answer accuracy stays on par with baselines yet evidence localization improves.
Despite relying solely on a pretrained Qwen2 backbone with no task‑specific fine‑tuning, MUPA outperforms specialized architectures in both Acc@GQA and grounding metrics. By preserving the native QA capability of Qwen2 and enhancing the accuracy of visual evidence retrieval, our framework produces more trustworthy predictions grounded firmly in video content.

\begin{figure}[t]
  \centering
  \includegraphics[width=1\textwidth]{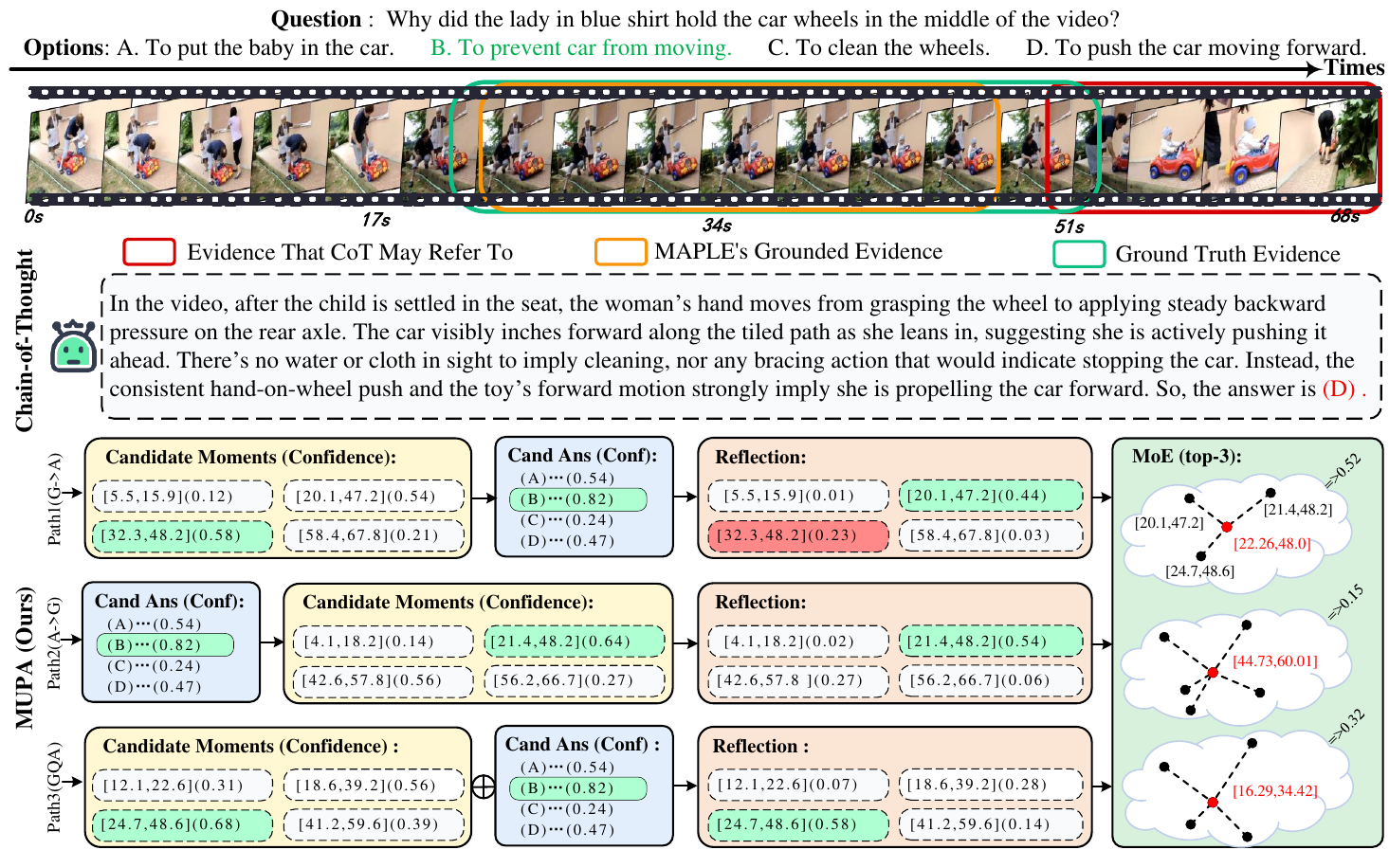} 
  \vspace{-12pt}
  \caption{Visualization of the MUPA workflow. Each of the three reasoning paths produces candidate answer and evidence. Reflection Agent then scores and aggregates the fused outputs to select the final answer and top-\(k\) temporal evidence spans.}
  \label{fig:fig4}
  \vspace{-6pt}
\end{figure}

\subsection{Qualitative Analysis}
In Figure.~\ref{fig:fig4}, we provide a detailed visualization of MUPA's multi-path reasoning process, demonstrating how MUPA runs three paths and reflects to progressively derive the correct answer while avoiding potential mistakes or confusion. This concise case study highlights MUPA’s strength in filtering misleading signals and unifying the candidates of complementary reasoning paths into an accurate, grounded prediction.
\begin{table}[!htbp]
\vspace{-5pt}
\centering
\caption{Ablations to study the effects of multi-path and reflection.}
\small
\begin{tabular}{c|cc|cc|cc|cc}
\toprule
\multirow{2}{*}{Reasoning Path} 
  & \multicolumn{2}{c|}{Reflection} 
  & \multicolumn{2}{c|}{IoU} 
  & \multicolumn{2}{c|}{IoP} 
  & \multirow{2}{*}{Acc@QA} 
  & \multirow{2}{*}{Acc@GQA} \\
\cmidrule(lr){2-3}\cmidrule(lr){4-5}\cmidrule(lr){6-7}
& w & w/o & R1@0.5 & mIoU & R1@0.5 & mIoP &  &  \\
\midrule
Single path-1 (G$\to$A) & \xmark & \checkmark & 22.4 & 25.2 & 32.2 & 35.1 & 70.4 & 24.8 \\
Single path-1 (G$\to$A) & \checkmark & \xmark & 23.1 & 27.6 & 35.7 & 37.4 & 71.0 & 26.4 \\
Single path-2 (A$\to$G) & \xmark & \checkmark & 21.2 & 25.9 & 32.6 & 35.9 & 69.8 & 25.8\\
Single path-2 (A$\to$G) & \checkmark & \xmark & 22.9 & 26.9 & 37.1 & 37.3 & 70.8 & 27.4 \\
Single path-3 (GQA)     & \xmark & \checkmark & 23.2 & 27.4 & 35.7 & 38.8 & 70.6 &  26.6\\
Single path-3 (GQA)     & \checkmark & \xmark & 23.8 & 29.2 & 37.2 & 38.2 & 70.8 & 27.6 \\
\midrule
\rowcolor[gray]{0.9}
MUPA-2B (Multi-path)     & \checkmark & \xmark & \textbf{25.6} & \textbf{27.2} & \textbf{38.7} & \textbf{39.1} & \textbf{72.8} & \textbf{28.7} \\
\bottomrule
\end{tabular}
\label{tab:ablation_test}
\vspace{-15pt}
\end{table}

\subsection{Ablation Studies}

\noindent\textbf{What Effects does Multi-path Contribute?}
The ablation results shown in Table~\ref{tab:ablation_test} confirm that MUPA’s multi-path design is a principal factor in its performance. Specifically, when leveraging both the $G \rightarrow A$ and $A \rightarrow G$ reasoning paths, the model achieves more robust grounding and answer accuracy than any single-path variant. Each path offers complementary strengths: the $G \rightarrow A$ path grounds the question before proposing an answer, while the $A \rightarrow G$ path verifies candidate answers against temporal evidence. Variants restricted to a single path exhibit declines in grounding fidelity and correctness, demonstrating that neither path alone fully captures the nuances of grounded VideoQA. By integrating these parallel reasoning trajectories, MUPA synthesizes complementary signals, substantially improving alignment precision and answer reliability.

\noindent\textbf{Does the Reflection Agent Boost Grounding Fidelity and Answer Accuracy?}
The ablation study further demonstrates that the Reflection Agent enhances both answer accuracy and grounding fidelity, as reported in Table~\ref{tab:ablation_test}. When enabled, this component engages in an iterative verification process that critiques initial predictions and reinforces consistency between answers and evidence. Through self-correction, the Reflection Agent identifies unsupported or erroneous outputs and refines them, leading to final responses that are both more accurate and more faithfully grounded in video segments. In contrast, disabling the Reflection Agent causes the model to adhere to its initial outputs even when they lack sufficient justification. These observations affirm the Reflection Agent’s crucial role in strengthening MUPA’s reliability and grounding quality.

\section{Conclusion}
In this work, we introduce MUPA, a principled multi-agent collaborative and multi-path reasoning framework that tackles Grounded VideoQA tasks. Through four collaborative agents and a reflection mechanism that verifies intra-path consistency via PoE and fuses inter-path hypotheses through MoE, MUPA narrows the long-standing gap between answer accuracy and evidence fidelity. Empirically, MUPA consistently improves grounding and grounded QA, achieving SOTA performance on different Grounded VideoQA benchmarks.


\appendix
\label{appendix}
\input{Supplymentary_materials_body}

\end{document}

%% file: Supplymentary_materials_body.tex
 


\newpage

\section{Discussion on Entropy-based Measurement}

In this document, we provide more descriptions of MUPA, including its implementation details and datasets used for training. Additional experiments and prompt templates are also incorporated. Finally, we discuss the limitations of our approach and outline directions for future work.

\section{Model Details}

\subsection{Implementation Details}

We adopt the 2B and 7B variants of Qwen2-VL~\cite{wang2024qwen2} as our backbone models. Each component (Grounder, GQA, and Verifier) is fine-tuned separately on its respective dataset for one epoch, using a global batch size of 32. Optimization is performed with AdamW, where the learning rate is set to $1\times10^{-4}$ for the Grounder and GQA modules and $5\times10^{-5}$ for the Verifier. A linear warmup schedule is applied over the first 3\% of total training steps. At inference, redundant spans generated by the Grounder are eliminated via non-maximum suppression with an IoU threshold of 0.75. All experiments were run on 8 $\times$ NVIDIA H100 GPUs (80 GB each). Table~\ref{tab:hyper_para} summarizes the key hyper-parameters.

\begin{table}[!htbp]
  \centering
  \caption{Key hyper‑parameters for agents.}
  \label{tab:hyper_para}
  \begin{tabular}{lcccccc}
    \toprule
    \textbf{Role}  & \textbf{Max Tokens} & \textbf{Max \#Frames} & \textbf{FPS} \\
    \midrule
    Grounder       & 64  & 150 & 1 \\
    GQA            & 64  & 150 & 1 \\
    Verifier       & 64  & 64  & 2 \\
    Answerer       & 256 & 32  & 2 \\
    \bottomrule
  \end{tabular}
\end{table}

\begin{figure}[!htbp]
  \centering
\includegraphics[width=\textwidth]{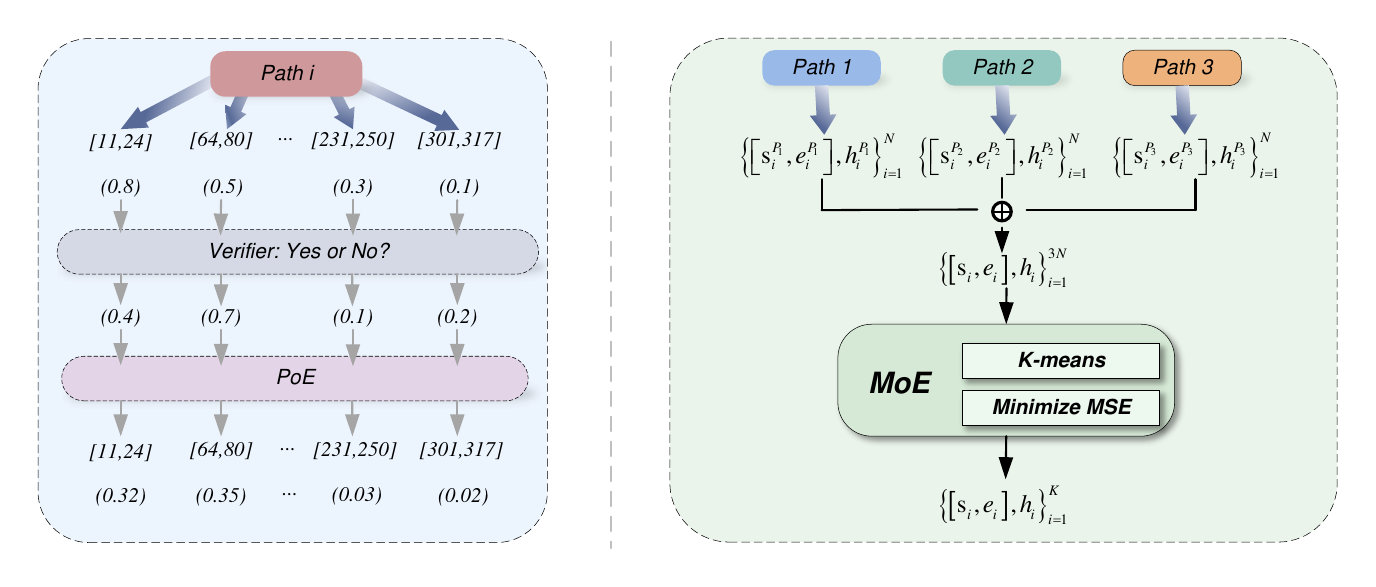} 
\vspace{-20pt}
  \caption{Illustration of the Reflection agent’s two core operations: \textbf{(Left)} Single‐path verification of candidate moments using a binary Verifier followed by PoE. \textbf{(Right)} Multi‐path fusion via a MoE module that performs K‐means clustering and MSE minimization to yield the final top‐$K$ fused moment predictions.}
  \label{fig:sup_fig1}
\end{figure}

\subsection{More Details of Reflection Agent}
 
In this section, we complement Section 3.3 of the main paper with detailed workflow diagrams shown in Fig.~\ref{fig:sup_fig1} and implementation specifics, including pseudocode and parameter settings for the two core operations of the Reflection Agent.

\begin{algorithm}[H]
\caption{Single‐Path Verification and PoE Re‐scoring}
\KwIn{Grounder spans $\{(S_{ik},c_{ik})\}_{k=1}^N$, question $q$}
\KwOut{Refined span $\hat S_i$, path confidence $p_i$}
Insert \texttt{<SEG\_START>}/\texttt{<SEG\_END>} and zoom‐in each $S_{ik}$ by 50\%\;
$(\ell_{\mathrm{yes}},\ell_{\mathrm{no}})\leftarrow \mathrm{Verifier}(S_{ik},q)$\;
$v_{ik}\leftarrow\sigma(\ell_{\mathrm{yes}}-\ell_{\mathrm{no}})$\;
$p_{ik}\leftarrow c_{ik}\,v_{ik}$\;
$k^*\leftarrow\arg\max_k p_{ik}$\;
$\hat S_i\leftarrow S_{ik^*}$,\quad $p_i\leftarrow p_{ik^*}$\;
\end{algorithm}

\begin{algorithm}[H]
\caption{Multi‐Path MoE Fusion}
\KwIn{Verified spans $\{(S_{ik},p_{ik})\}$ and path predictions $\{A_i,p_i\}_{i=1}^M$}
\KwOut{Final answer $\hat A$, top‐$K$ spans $\{\tilde S_j\}_{j=1}^K$}
$\displaystyle\hat A\leftarrow\arg\max_a\sum_i p_i\,\mathbf1[A_i=a]$\;
Normalize weights: $w_{ik}\leftarrow p_{ik}/\sum_{j,\ell}p_{j\ell}$\;
Perform weighted $K$-means clustering on $\{\mathbf x_{ik}=(s_{ik},e_{ik})\}$ with weights $w_{ik}$\;
\For{$j\leftarrow1$ \KwTo $K$}{
  $I_j\leftarrow\{(i,k)\mid c(ik)=j\}$\;
  $\tilde{\mathbf C}_j\leftarrow\sum_{(i,k)\in I_j}w_{ik}\,\mathbf x_{ik}\big/\sum_{(i,k)\in I_j}w_{ik}$\;
  $\tilde S_j\leftarrow(\tilde s_j,\tilde e_j)$\;
}
\Return{$\hat A$, $\{\tilde S_j\}$}\;
\end{algorithm}

\textbf{Parameter Settings.} Below we list and explain each hyperparameter, referring to the defaults in our implementation and the values used in our experiments:
\begin{itemize}
  \item $N=5$: the number of candidate spans per reasoning path produced by the Grounder (default in our Verifier loop).  
  \item $K=5$: the number of clusters (final evidence segments) in the Mixture‐of‐Experts fusion. Although our \texttt{weighted\_kmeans} implementation defaults to \texttt{n\_clusters=5}, we override this to 3 to focus on the top‐3 moments.  
  \item \textbf{K‐means max iterations} $=10$: the \texttt{max\_iters} argument in \texttt{weighted\_kmeans}, limiting the number of EM updates (default also 10).  
  \item \textbf{Convergence tolerance} $\varepsilon=10^{-6}$: the \texttt{eps} parameter in \texttt{weighted\_kmeans}, used to avoid degenerate cluster assignments and to check for centroid stability (default also 10)
\end{itemize}

\section{Experimental Details}

\subsection{Training Datasets}

This section outlines the datasets used to train each component of MUPA. We pretrain on a diverse collection of large-scale video-grounded QA and moment retrieval benchmarks to instill robust localization and reasoning capabilities, and then fine-tune each module on task-specific splits to adapt to our evaluation protocols.

\begin{table}[!htbp]
  \centering
  \caption{Supervised fine-tuning datasets for MUPA}
  \label{tab:pretrain-datasets}
  \begin{tabularx}{\textwidth}{l c X}
    \toprule
    \textbf{Role}  & \textbf{\#Samples} & \textbf{Pretraining Datasets} \\
    \midrule
    Grounder  & 210K & QVHighlights (5K), DiDeMo (33K), TACoS (9K), QuerYD (19K), HiREST\textsubscript{mr} (8K), HiREST\textsubscript{step} (4K), CosMo-Cap (87K), InternVid-VTime (54K) \\
    \midrule
    GQA       & 179K & QVHighlights-QA (5K), TACoS-QA (9K), CosMo-Cap-QA (87K), DeVE-QA (78K) \\
    \midrule
    Verifier  & 232K & DiDeMo-Verify (165K), TACoS-Verify (43K), QVHighlights-Verify (24K) \\
    \bottomrule
  \end{tabularx}
\end{table}

\textbf{Dataset for GQA.} The QVHighlights-QA, TACoS-QA, and CosMo-Cap-QA datasets are derived from the original QVHighlights~\cite{lei2107qvhighlights}, TACoS~\cite{regneri2013grounding}, and CosMo-Cap~\cite{wang2024cosmo} benchmarks, respectively.  We employ GPT-4o-mini~\cite{hurst2024gpt} as a prior-knowledge assistant to generate question–answer pairs from the existing video descriptions. Then we apply a BlindQA filtering step, where candidate answers are constrained to 4–6 choices and the model answers without access to video content, so that the resulting BlindQA accuracy is approximately 20\% at chance level. Here, “BlindQA” refers to a baseline setting in which answers are produced solely from the question text, serving both to assess dataset bias and to quantify the performance gain when incorporating visual input.

\textbf{Datasets for Verifier.} Verifier datasets are constructed by running the pretrained Grounder on each source benchmark and collecting its predicted spans as pseudo-ground-truth.  In particular, the “mr” and “step” suffixes denote the moment-retrieval and step-localization subsets of HiREST~\cite{hiREST2025}, respectively.  This process yields DiDeMo-Verify, TACoS-Verify, and QVHighlights-Verify, each of which reflects the Grounder’s hypotheses and provides a realistic training signal for the Verifier module.

\subsection{More Comparisons with SOTA}

In this section, we evaluate MUPA on three additional Moment Retrieval datasets, including TACoS~\cite{regneri2013grounding}, ActivityNet-Captions~\cite{krishna2017dense}, and ActivityNet-RTL~\cite{huang2024lita}. These datasets cover a wide range of domains (e.g., instructional videos, cooking scenarios, and diverse open-domain activities) and query formats, providing a comprehensive testbed for zero-shot or fine-tuned evaluation of our framework's temporal reasoning capabilities.
Moment Retrieval focuses on localizing text queries in time. We measure performance by R@$\{0.3,0.5,0.7\}$ (Recall at various IoU thresholds) and mIoU (mean IoU). In addition, certain benchmarks report P@0.5 (Precision at IoU$\,\ge 0.5$) and mAP (mean average precision) as complementary metrics for retrieval quality.


\begin{table}[!htbp]
\centering
\caption{Fine-tuned video temporal grounding on TACoS~\cite{regneri2013grounding}. \textbf{FT} indicates whether fine-tuned on the downstream training set. MUPA demonstrates strong generalization, with its zero-shot scores outperforming all zero-shot baseline and surpassing fine-tuned variants. }
\small
\begin{tabular}{l |cc| c c c c}
\toprule
Method & Size & FT & R1@0.3 & R1@0.5 & R1@0.7 & mIoU \\
\midrule
\multicolumn{7}{l}{\textcolor{gray}{\emph{Non‑LLM‑based Specialists}}} \\ 
2D-TAN\cite{zhang2020learning} & -- & \cmark & 40.0 & 28.0 & 12.9 & 27.2 \\
VSLNet\cite{zhang2020span} & -- & \cmark & 35.5 & 23.5 & 13.1 & 25.0 \\
Moment-DETR\cite{lei2107qvhighlights} & -- & \cmark & 38.0 & 24.7 & 12.0 & 25.5 \\
UniVTG\cite{lin2023univtg} & -- & \cmark & 51.4 & 35.0 & 17.4 & 33.6 \\
R2-Tuning\cite{liu2024r} & -- & \cmark & 49.7 & 38.7 & 25.1 & 35.9 \\
\midrule
\multicolumn{7}{l}{\textcolor{gray}{\emph{LLM‑based Models}}} \\ 
VideoMind-2B\cite{liu2025videomind} & 2B & \cmark & 38.6 & 26.9 & 15.5 & 27.4 \\
VideoMind-7B\cite{liu2025videomind} & 7B & \cmark & \textbf{49.5} & 36.2 & 21.4 & 34.4 \\
\midrule
\rowcolor[gray]{0.9} MUPA-2B (Ours) & 2B & \cmark & 41.9 & 37.1 & 29.5 & 34.2 \\
\rowcolor[gray]{0.9} MUPA-7B (Ours) & 7B & \cmark & 46.1 & \textbf{40.9} & \textbf{32.6} & \textbf{37.8} \\
\bottomrule
\end{tabular}
\label{tab:tacos}
\end{table}

\textbf{TACoS.} In Table~\ref{tab:tacos}, our MUPA achieves a mIoU of 37.8\% on the cooking-oriented TACoS dataset~\cite{regneri2013grounding}, edging out the best specialist by a comfortable margin. These gains highlight MUPA's ability to track fine temporal boundaries in long, compositional activities.

\section{Prompt Templates}

We present the prompts used in this work, including the input prompts for each agent of MUPA.

\begin{grounderbox}
You are acting as the grounder now. Given a video and a text query, your goal is to temporally localize the video moment described by the query. If the query is directly describing a moment, simply localize it according to its content. Otherwise, if the moment is described as 'before/after a pivotal event', you need to determine the actual event it refers to. The localized moment should only cover the target event. Now I give you the query: '{}'. Please think carefully and provide your response.
\end{grounderbox}

\begin{gqabox}
You are acting as the GQA Agent now.
Given a video and a multiple-choice question, you have two tasks:
1) Trigger the video-moment retrieve pipeline to temporally localize the video moment described by the question by generating exactly \texttt{<REG\_TOKEN>}.
2) Choose the best answer from given options.

Question: {}

Options:
{}

Please reply exactly in this format:
1) The relevant moment happens in \texttt{<REG\_TOKEN>}
2) Best choice: <Option>
\end{gqabox}

\begin{verifierbox}
You are acting as the verifier now. You will be presented a text query describing a moment that potentially happens in the given video. Your task is to identify whether the video segment between \texttt{<SEG\_S\_TOKEN>} and \texttt{<SEG\_E\_TOKEN>} perfectly covers the moment. If the described moment can be seen in the video, please focus on verifying whether the moment starts at \texttt{<SEG\_S\_TOKEN>} and ends at \texttt{<SEG\_E\_TOKEN>}. Respond with 'Yes' if you think the moment boundaries are correct, otherwise 'No'. If the described moment cannot be seen in the video, respond with 'No' directly. Now I give you the query: '{}'. Please think carefully and respond with 'Yes' or 'No' directly.
\end{verifierbox}

\section{Limitations and Future Work}

We have to admit that MUPA incurs higher inference costs because each query is processed by several paths, and it currently lacks a planner to adaptively prune redundant trajectories. Future work will explore dynamic path selection, incorporate spatial grounding beyond temporal spans, and extend the framework to broader vision–language tasks such as instructional video understanding and multi-step reasoning.
Nevertheless, our study confirms that combining multi-multipath reasoning with reflection boosts answer reliability and evidence alignment, positioning MUPA as an important step toward Trustworthy Multimodal Large Language Models.
